OXFORD

Data and text mining

# CoxKAN: Kolmogorov-Arnold networks for interpretable, high-performance survival analysis

**William Knottenbelt[1,2,‡], William McGough[1,3,‡], Rebecca Wray[1,3], Woody Zhidong Zhang[1,3], Jiashuai Liu[4], Ines Prata Machado[1,3], Zeyu Gao[1,3,*], Mireia Crispin-Ortuzar[1,3,*]**

[1]Department of Oncology, University of Cambridge, Cambridge, CB2 0XZ, United Kingdom
[2]Department of Physics, University of Cambridge, Cambridge, CB3 0HE, United Kingdom
[3]CRUK Cambridge Centre, University of Cambridge, Cambridge, CB2 0RE, United Kingdom
[4]School of Computer Science, Xi'an Jiaotong University, Xi'an, 710049, China

*Corresponding authors. Zeyu Gao, Department of Oncology, University of Cambridge, Adrian Way, Cambridge CB2 0XZ, United Kingdom. E-mail: zg323@cam.ac.uk; Mireia Crispin-Ortuzar, Department of Oncology, University of Cambridge, Cambridge, CB2 0XZ, United Kingdom. E-mail: mc973@cam.ac.uk

‡= equal contribution.

## Abstract

**Motivation:** Survival analysis is a branch of statistics that is crucial in medicine for modeling the time to critical events such as death or relapse, in order to improve treatment strategies and patient outcomes. Selecting survival models often involves a trade-off between performance and interpretability; deep learning models offer high performance but lack the transparency of more traditional approaches. This poses a significant issue in medicine, where practitioners are reluctant to use black-box models for critical patient decisions.

**Results:** We introduce CoxKAN, a Cox proportional hazards Kolmogorov-Arnold Network for interpretable, high-performance survival analysis. Kolmogorov-Arnold Networks (KANs) were recently proposed as an interpretable and accurate alternative to multi-layer perceptrons. We evaluated CoxKAN on four synthetic and nine real datasets, including five cohorts with clinical data and four with genomics biomarkers. In synthetic experiments, CoxKAN accurately recovered interpretable hazard function formulae and excelled in automatic feature selection. Evaluations on real datasets showed that CoxKAN consistently outperformed the traditional Cox proportional hazards model (by up to 4% in C-index) and matched or surpassed the performance of deep learning-based models. Importantly, CoxKAN revealed complex interactions between predictor variables and uncovered symbolic formulae, which are key capabilities that other survival analysis methods lack, to provide clear insights into the impact of key biomarkers on patient risk.

**Availability and implementation:** CoxKAN is available at GitHub and Zenodo.

## 1 Introduction

Survival analysis, also known as time-to-event analysis, models the timing of significant events across various fields, including medicine, engineering, economics, and insurance (Wang *et al.* 2019). In medicine, particularly in oncology, survival analysis is critically important for modeling events such as death and relapse. It plays a key role in identifying biomarkers and prognostic factors (Ou *et al.* 2021, Saegusa *et al.* 2021), assessing treatment efficacy (Monnickendam *et al.* 2019, Le-Rademacher and Wang 2021), and developing personalized treatment plans (Katzman *et al.* 2018).

The most prevalent survival model is the Cox proportional hazards model (CoxPH) (Cox 1972), which posits a linear relationship between a patient's covariates (e.g. age, blood pressure) and the *log-partial hazard*—a measure of the patient's risk of experiencing an event. This model offers clear interpretability, as it allows direct observation of how each covariate influences risk. However, its linear assumption can be overly simplistic, often introducing significant bias. In contrast, machine learning-based methods, including deep learning, typically allow a non-linear relationship between covariates and risk, potentially enhancing performance. These methods range from random survival forests (Ishwaran and Kogalur 2007, Ishwaran *et al.* 2008) and Bayesian models utilizing Gaussian processes (Fernandez *et al.* 2016, Vaswani *et al.* 2017) to dependent logistic regression (Yu *et al.* 2011), and extend to powerful deep learning models like "DeepSurv" (Katzman *et al.* 2018), and "Cox-nnet" (Ching *et al.* 2018). Elsewhere, non-log-partial hazard deep learning approaches predict survival times directly, without the partial hazard assumptions of Cox methods, and achieve comparable performance (Lee *et al.* 2018, Kvamme *et al.* 2019, Rindt *et al.* 2022).

Deep learning models are versatile and have been used extensively for survival analysis, achieving state-of-the-art performance across numerous datasets in many domains (Kim *et al.* 2020, Kvamme and Borgan 2021, Nagpal *et al.* 2021, Rindt *et al.* 2022). However, the increased flexibility of these models often comes at the expense of interpretability. Multilayer perceptrons (MLPs) are the fundamental building blocks of deep learning architectures and consist of multiple layers of neurons that process aspects of the data progressively to capture complex patterns. Despite their power, MLPs are not inherently interpretable, which is a challenge for clinical adoption. The search for more interpretable machine learning techniques remains an active area of research (Langbein *et al.* 2024, Wiegrebe *et al.* 2024).

Kolmogorov-Arnold Networks (KANs) (Liu *et al.* 2024) have recently been introduced as an alternative to MLPs, offering enhanced interpretability and accuracy. Unlike MLPs, which rely on linear weights, KANs utilize learnable activation functions on the edges of the network and sum these functions at the nodes ("neurons"). These activation functions are parameterized as B-spline curves with learnable coefficients, enabling them to approximate any univariate function effectively. As a result, KANs can solve complex problems using very few hidden neurons, which simplifies interpretability. This is further enhanced by fitting symbolic operators to the learned activation functions, ultimately leaving a symbolic formula in place of the network. Initially introduced in physics, KANs have since found multiple applications in various fields, including time series analysis (Vaca-Rubio *et al.* 2024), medical image segmentation (Li *et al.* 2024), and satellite image classification (Cheon 2024). Despite their various advantages, original KANs are not ideally suited for direct application in survival analysis due to inherent challenges. Primarily, the slow training process of KANs makes them impractical when dealing with large, high-dimensional survival datasets. Additionally, survival datasets may involve intricate relationships and interactions that cannot be appropriately captured by the default symbolic fitting mechanism of the original KANs.

In this work, we introduce CoxKAN, the first KAN-based framework for interpretable survival analysis. CoxKAN addresses these challenges by implementing a fast approximation to the Cox loss, reducing training times, and making it more feasible to handle high-dimensional data in survival analysis. Furthermore, CoxKAN leverages activation function pruning for automatic feature selection, effectively streamlining the analysis of complex, high-dimensional datasets. Additionally, CoxKAN incorporates a progressive symbolic regression involving PySR (Cranmer 2023), enabling more controlled and effective management of the bias-variance tradeoff to produce interpretable symbolic representations. The key contributions of our work with CoxKAN are delineated as follows: (i) We have adapted Kolmogorov-Arnold Networks (KAN) for survival analysis, enhancing their applicability and effectiveness for this field; (ii) CoxKAN has been rigorously tested with synthetic data, proving its ability to accurately derive symbolic formulae for the hazard function under complex and noisy conditions; (iii) Extensive testing on five clinical datasets demonstrates that CoxKAN outperforms traditional Cox models and matches or exceeds DeepSurv performance, while effectively identifying key biomarkers and complex variable interactions; (iv) Evaluation of four high-dimensional genomics datasets shows that CoxKAN effectively handles complex data distributions, aiding oncologists in understanding cancer biology intricacies; (v) CoxKAN is released as an open-source Python toolkit.

# 2 Methods

## 2.1 Survival analysis

Survival time is commonly described by the survival function $S(t) = P(T \geq t)$, which represents the probability that a patient survives beyond time $t$, and the hazard function $h(t)$, which is the instantaneous event rate at time $t$ given survival until $t$. Specifically, the hazard function is defined as $h(t) = \lim_{\Delta t \to 0} \frac{P(t \leq T < t+\Delta t | T \geq t)}{\Delta t}$, and the probability density function $f(t)$ can be expressed as $f(t) = h(t)S(t)$. The survival function can further be related to the hazard function by $S(t) = \exp\left(-\int_0^t h(s)ds\right)$.

Survival data for a patient includes (i) covariates $\mathbf{x}$ (predictor variables), (ii) time duration $t$, and (iii) an event indicator $\delta$, where $\delta = 1$ if the event is observed and $\delta = 0$ for right-censored cases. In cases of right-censoring, $t$ is the time until the last known contact.

### 2.1.1 Cox proportional hazards model (CoxPH)

A proportional hazards model is one that assumes the hazard function is given by $h(t, \mathbf{x}) = h_0(t) \cdot \exp(\theta(\mathbf{x}))$, where $h_0(t)$ represents the baseline hazard, and $\theta(\mathbf{x})$ is the log-partial hazard, which reflects overall patient risk independently of time. The Cox proportional hazards model (CoxPH) (Cox 1972), a widely used survival regression model, defines the log-partial hazard as a linear combination of covariates: $\widehat{\theta}_{CPH}(\mathbf{x}) = \boldsymbol{\beta}^T \mathbf{x} = (\beta_1 x_1 + \beta_2 x_2 + \cdots + \beta_n x_n)$. Given a dataset of $N$ patients, $\{(\mathbf{x}_i, t_i, \delta_i)\}_{i=1}^N$, the weights $\boldsymbol{\beta}$ are optimized using the Cox partial likelihood, $\prod_{i:\delta_i=1} \frac{\exp(\theta(\mathbf{x}_i))}{\sum_{j \in \mathcal{R}(t_i)} \exp(\theta(\mathbf{x}_j))}$, where $\mathcal{R}(t_i)$ is the "risk set" containing all patients still alive at time $t_i$.

### 2.1.2 DeepSurv

DeepSurv is a proportional hazards model that leverages a neural network to predict the log-partial hazard (Faraggi and Simon 1995, Katzman *et al.* 2018). DeepSurv is trained using the Cox loss function, defined as the negative log of the partial likelihood, $-\sum_{i:\delta_i=1} [\widehat{\theta}(\mathbf{x}_i) - \log(\sum_{j \in \mathcal{R}(t_i)} \exp(\widehat{\theta}(\mathbf{x}_j)))]$.

### 2.1.3 Cox-nnet

Similar to DeepSurv, Cox-nnet is a proportional hazards model that approximates the log partial hazard with a neural network (Ching *et al.* 2018). The only difference between Cox-nnet and DeepSurv is the restriction of Cox-nnet to one hidden layer, whereas DeepSurv is let to be arbitrarily deep.

### 2.1.4 SuMo

Unlike Cox-nnet, DeepSurv, or CoxPH, SuMo drops the partial-hazard assumption and learns the risk-time distribution profile directly, allowing direct prediction of each individual's survival curve (Rindt *et al.* 2022). Similarly to Cox-nnet, SuMo consists of a small neural network (between 1–3 hidden layers). SuMo enforces risk-time monoticity in its architectural design, and uses a negative log-likelihood loss for right-censored survival data $-\frac{1}{N}(\sum_{i:\delta_i=0} \log(\max(S_i(t_i), \varepsilon)) + \sum_{i:\delta_i \neq 0} \log(\max(\lambda_i(t_i), \varepsilon)))$ where $S_i$ is the predicted survival probability for sample $i$ at time $t_i$, $\lambda_i$ is the gradient of $S$ at $t_i$, and $\varepsilon$ is a small constant.

## 2.2 Kolmogorov-Arnold networks

Kolmogorov-Arnold Networks (KANs) (Liu *et al.* 2024) are a class of neural networks inspired by the Kolmogorov-Arnold representation theorem, which states that any continuous function can be represented as a finite composition of univariate functions. Structured similarly to Multi-Layer Perceptrons (MLPs), KANs consist of consecutive layers of neurons, each depending on the previous one. The shape of a KAN is defined by $[n_0, n_1, \ldots, n_L]$, with $n_l$ representing the number of neurons in the $l^{\text{th}}$ layer, $\mathbf{x}_l = (x_{l,1}, x_{l,2}, \ldots, x_{l,n_l})^T$. Here, $\mathbf{x}_0$ is the input to the network and $\mathbf{x}_L$ is the final output. Unlike MLPs, KANs use $n_l \cdot n_{l+1}$ learnable activation functions between each layer $l$ and $l+1$, parameterized using B-splines, which can approximate arbitrary univariate functions. The activation function connecting the $i^{\text{th}}$ neuron in the $l^{\text{th}}$ layer to the $j^{\text{th}}$ neuron in the $(l+1)^{\text{th}}$ layer is denoted $\phi_{l,j,i}$. The output of $(l+1)^{\text{th}}$ layer is computed as:

$$x_{l+1,j} = \sum_{i=1}^{n_l} \phi_{l,j,i}(x_{l,i}), \quad j = 1, \ldots, n_{l+1} \tag{1}$$

which can be expressed in matrix form as:

$$\mathbf{x}_{l+1} = \underbrace{\begin{pmatrix} \phi_{l,1,1}(\cdot) & \phi_{l,1,2}(\cdot) & \cdots & \phi_{l,1,n_l}(\cdot) \\ \phi_{l,2,1}(\cdot) & \phi_{l,2,2}(\cdot) & \cdots & \phi_{l,2,n_l}(\cdot) \\ \vdots & \vdots & & \vdots \\ \phi_{l,n_{l+1},1}(\cdot) & \phi_{l,n_{l+1},2}(\cdot) & \cdots & \phi_{l,n_{l+1},n_l}(\cdot) \end{pmatrix}}_{\mathbf{\Phi}_l} \mathbf{x}_l, \tag{2}$$

The network's output for an input vector $\mathbf{x} \in \mathbb{R}^{n_0}$ can then be described as:

$$\text{KAN}(\mathbf{x}) = (\mathbf{\Phi}_{L-1} \circ \mathbf{\Phi}_{L-2} \circ \cdots \circ \mathbf{\Phi}_1 \circ \mathbf{\Phi}_0)\mathbf{x}. \tag{3}$$

All operations in KANs are differentiable, enabling training via backpropagation.

#### 2.2.1 Activation functions

The activation function $\phi(x)$ in KANs is defined as:

$$\phi(x) = w_b b(x) + w_s \text{spline}(x), \tag{4}$$

where $w_b$ and $w_s$ are trainable weights that control the magnitude of the activation, and $b(x)$ is a non-trainable basis function for stability (analogous to a residual connection). The spline function is defined as $\text{spline}(x) = \sum_{i=0}^{G+k-1} c_i B_{i,k}(x)$, where the $c_i$ are trainable parameters and the $B_{i,k}(\cdot)$ are B-spline basis functions of degree $k$ on $G$ grid intervals. In this work, we use $k=3$, $G \in \{3,4,5\}$, and $b(x)$ as either $x$ or $\text{silu}(x) = \frac{x}{1+e^{-x}}$. It can be shown that for sufficiently high $k$ and $G$, $\text{spline}(x)$ can approximate any smooth uni-variate function defined on a bounded domain with arbitrary accuracy.

#### 2.2.2 Regularization

For efficiency and interpretability, KAN should be kept as small and simple as possible. However, determining the optimal structure in advance can be challenging. To address this, (Liu *et al.* 2024) introduced a regularization and pruning scheme that simplifies a KAN starting from a larger network. This involves adding regularization terms to the loss function to promote sparsity among KAN neurons and spline coefficients.

The L1 norm of an activation function $\phi$ is defined to be its average magnitude over the training batch of $N_B$ inputs:

$$|\phi|_1 \equiv \frac{1}{N_B} \sum_{s=1}^{N_B} |\phi(x^{(s)})|, \tag{5}$$

and that of its spline coefficients $\mathbf{c}$ is $|\mathbf{c}|_1 = \frac{1}{G+K} \sum_{i=0}^{G+k-1} |c_i|$.

Then, the L1 norm of a full KAN layer $\mathbf{\Phi}$ with $n_{\text{in}}$ inputs and $n_{\text{out}}$ outputs, is given by the sum of the L1 norms of the individual activations:

$$|\mathbf{\Phi}|_1 \equiv \sum_{i=1}^{n_{\text{in}}} \sum_{j=1}^{n_{\text{out}}} |\phi_{i,j}|_1. \tag{6}$$

Similarly, for the layer's collective set of spline coefficients $\mathbf{C}$, we have $|\mathbf{C}|_1 \equiv \sum_{i=1}^{n_{\text{in}}} \sum_{j=1}^{n_{\text{out}}} |\mathbf{c}_{i,j}|_1$. Next, the entropy of $\mathbf{\Phi}$ is defined as:

$$S(\mathbf{\Phi}) \equiv -\sum_{i=1}^{n_{\text{in}}} \sum_{j=1}^{n_{\text{out}}} \frac{|\phi_{i,j}|_1}{|\mathbf{\Phi}|_1} \log\left(\frac{|\phi_{i,j}|_1}{|\mathbf{\Phi}|_1}\right). \tag{7}$$

The overall regularization added to the loss function is formulated as follows:

$$R = \sum_{l=0}^{L-1} |\mathbf{\Phi}_l|_1 + \lambda_{ent} \sum_{l=0}^{L-1} S(\mathbf{\Phi}_l) + \lambda_{coef} \sum_{l=0}^{L-1} |\mathbf{C}_l|_1, \tag{8}$$

where $\lambda_{ent}, \lambda_{coef}$ are the weights controlling the entropy and coefficient regularization terms, respectively. This regularization on spline coefficients promotes simpler spline functions, reducing the risk of overfitting. Additionally, L1 and entropy regularization on activation magnitudes encourage sparsity in both activations (edges) and neurons (nodes) within the network. After this step, edges or nodes with an L1 norm below a specified threshold can be pruned, retaining only the most significant elements in the network.

### 2.3 CoxKAN

Building on KANs, we propose CoxKAN as a novel proportional hazards model, where a KAN structure is utilized to estimate the log-partial hazard through a single output node:

$$\hat{\theta}_{KAN}(\mathbf{x}) = \text{KAN}(\mathbf{x}). \tag{9}$$

The CoxKAN pipeline is shown in Fig. 1. First, the model is trained using a fast approximation of the Cox loss to ensure efficient training times. The second step involves pruning of activation functions, which enhances both interpretability and generalization by removing irrelevant features and simplifying the model. Finally, progressive symbolic fitting further strengthens interpretability and generalization by converting activation functions into simplified symbolic forms, making the model more transparent and robust. Similar to ablation, we report the test results of the CoxKAN pipeline following each step (training, pruning, and symbolic fitting) to ensure these steps improve performance, mirroring the original KAN paper (Liu *et al.* 2024).

#### 2.3.1 Training

CoxKAN is trained using the following objective:

$$\ell_{\text{total}} = \ell_{\text{Cox}} + \lambda R, \tag{10}$$

where, $R$ is the regularization term defined in Eq. (8), with $\lambda$ as the regularization weight, and $\ell_{\text{Cox}}$ is a fast approximation of the Cox loss (Kvamme *et al.* 2019). This approximation results in slight inaccuracies in the risk set $\mathcal{R}(t_i)$ when many patients share the same observed duration, but our experiments show that the performance of CoxKAN remains unaffected. Given that KANs are inherently slow to train, this approximation significantly speeds up training, making them practical for large-scale datasets without compromising accuracy. CoxKAN is optimized using Adam optimizer (Kingma and Ba 2015).

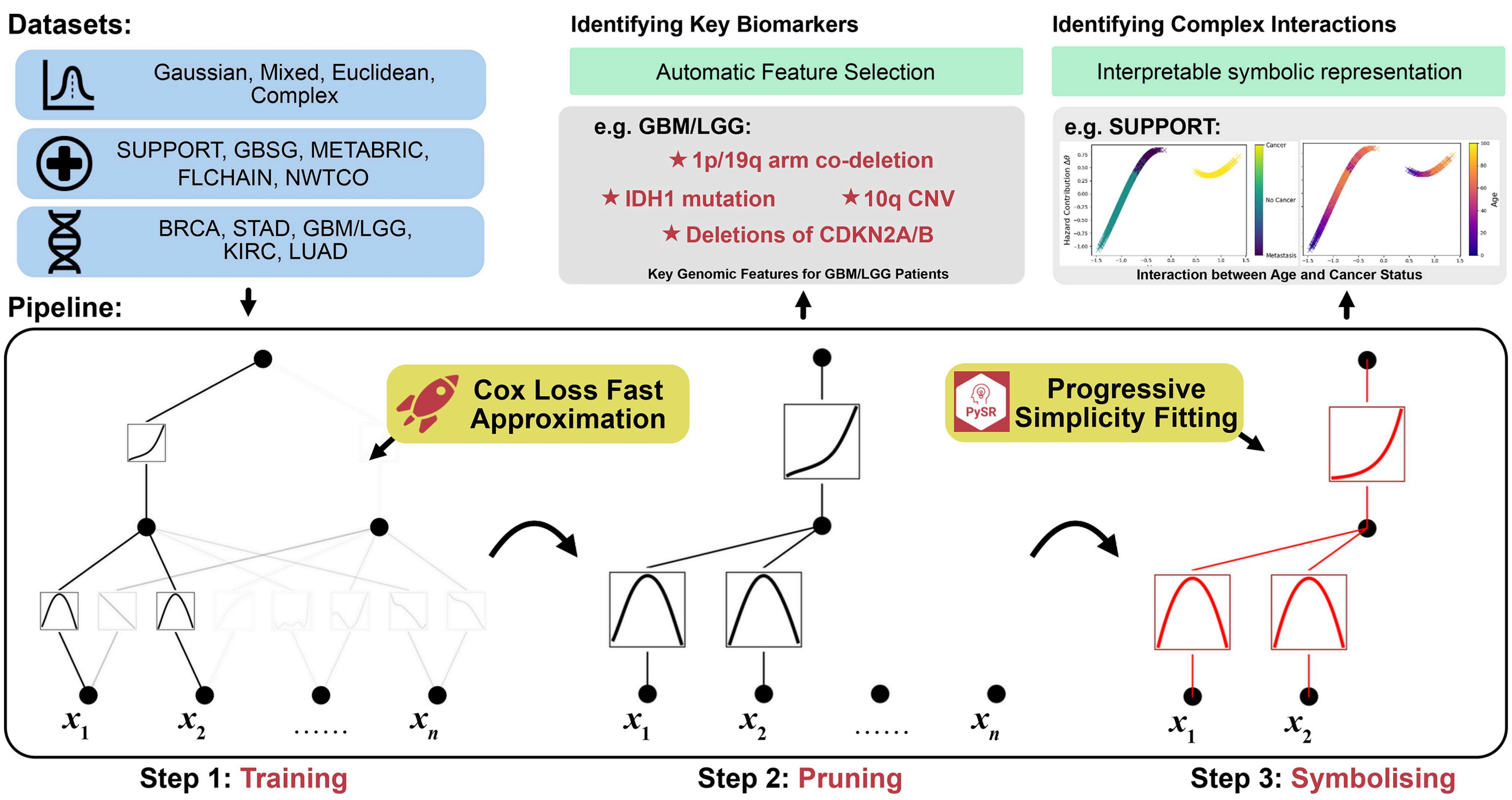

**Figure 1.** A study overview of how CoxKAN is trained and interpreted in this manuscript. CoxKAN is trained via a Cox-loss approximation on various datasets to predict each patient's partial hazard. Following training, network pruning, and symbolic-function fitting, insights emerge from inspection of the CoxKAN network structure. Key biomarkers are identified, and non-linear relationships between features and risk are characterized through the fitted symbolic expressions.

To efficiently explore the hyperparameter space, we implement random hyperparameter optimization with the Tree-structured Parzen Estimator (Watanabe 2023), optimizing the average C-Index over cross-validation on the training set. Finally, early stopping is applied based on the validation set C-Index.

Additionally, we use label encoding to handle categorical covariates in CoxKAN. The local control of B-splines allows different parameters to govern distinct regions of the input domain, enabling the network to manage each category independently, despite being encoded within the same dimension. In contrast, one-hot encoding would substantially increase the network's parameters, raising its risk of overfitting. To obtain a symbolic representation of the B-spline, we simply replace it with a discrete map.

### 2.3.2 Pruning

After training CoxKAN, we prune activation functions by removing those with L1 norms below a certain threshold, enabling automatic feature selection and refining the network's architecture. The L1 threshold for pruning is a tunable hyperparameter; however, when early stopping is applied using a validation set, we select the optimal L1 threshold based on validation performance. This pruning is especially advantageous in high-dimensional survival data, as it removes irrelevant features, reducing complexity and overfitting risk, and helping the model focus on the most relevant markers.

### 2.3.3 Symbolizing

As a crucial step in CoxKAN, *Symbolizing* enhances interpretability by converting activation functions into clear symbolic formulae rather than parameterized B-spline curves. In the original KAN approach (Liu *et al.* 2024), each activation function $\phi(x)$ is approximated by a known symbolic operator $f$ (e.g. sin or exp), expressed as $\phi(x) = cf(ax+b)+d$. The affine parameters $(a, b, c, d)$ are optimized by fitting the pre- and post-activation pairs $\{x^{(s)}, y^{(s)}\}_{s=1}^{M}$ through grid search and linear regression, with the fit quality assessed by $R^2$ (fraction of variance explained).

Here, we introduce *progressive symbolic fitting*, which begins with simple symbolic forms and increases in complexity as needed. This approach enhances interpretability by favoring simpler forms when possible, thus reducing model complexity and improving generalizability.

First, linear fits ($f(x) = ax+b$) are applied to each activation function and accepted if $R^2 > 0.99$. If rejected, CoxKAN applies `auto_symbolic` to select the best-fitting operator from a predefined library of 22 functions, as illustrated in Appendix, available as supplementary data at *Bioinformatics* online (Hyperparameters). Finally, if no suitable symbolic operator is identified, CoxKAN equipped with PySR (Cranmer 2023) determines a symbolic form for the activation function by using a genetic algorithm to perform symbolic regression on the pre- and post-activations $\{x^{(s)}, y^{(s)}\}_{s=1}^{M}$. This process, known as symbolic regression, searches a wide space of potential symbolic functions to find the most suitable approximation. It is particularly valuable for complex activation functions that are not neatly represented by a single symbolic operator, enabling CoxKAN to capture intricate data patterns while maintaining interpretability by favoring simpler forms whenever possible.

### 2.3.4 Remark

CoxKAN provides significant advantages over both CoxPH and the deep learning approaches tested in this manuscript: DeepSurv, Cox-nnet. and SuMo. Unlike CoxPH, which requires extensive manual feature engineering to capture non-linearities and interactions, CoxKAN learns these automatically. Although adding spline terms in CoxPH partially addresses non-linearity, it lacks the ability to capture interactions, introducing higher bias. In contrast to the deep learning approaches, CoxKAN provides transparent, interpretable expressions through symbolic fitting. This transparency is essential in survival analysis, where understanding the interactions and contributions of features is crucial for making informed predictions.

# 3 Materials

## 3.1 Evaluation on synthetic data

To ascertain whether CoxKAN can successfully recover symbolic formulae from survival data, we generated four synthetic datasets based on the proportional-hazards model, using custom symbolic formulae for the log-partial hazard, as shown in Appendix: Table 1, available as supplementary data at *Bioinformatics* online. The custom formulae include a Gaussian function; a "mixed" function with a linear combination of independent trigonometric and polynomial terms; a Euclidean distance function; and a complex function with multiple non-linearities (e.g. logarithmic and absolute values) and interactions across terms. We used a constant baseline hazard of 0.01, a uniform censoring distribution, and sampled the covariates uniformly in $[-1, 1]$ unless stated otherwise. We also added two irrelevant noisy covariates to each dataset. The data generation details can be found in the Appendix, available as supplementary data at *Bioinformatics* online (Synthetic Data Generation).

## 3.2 Evaluation on clinical data

To assess the real-world application of CoxKAN, we compare its performance on five clinical datasets to CoxPH and DeepSurv.

1) SUPPORT: The Study to Understand Prognoses and Preferences for Outcomes and Risks of Treatment (SUPPORT) dataset includes 7098 training and 1775 testing patients with serious illnesses. Covariates include age, race, comorbidities, diabetes, dementia, cancer status, vital signs (mean blood pressure, heart rate, respiration rate, temperature), serum sodium, white blood cell count, and serum creatinine (Knaus *et al.* 1995).
2) GBSG: Breast cancer data from the Rotterdam tumor bank (1546 training patients) and German Breast Cancer Study Group (686 testing patients) includes hormonal therapy, tumor size, menopausal status, age, lymph node count, progesterone receptor (PGR), and estrogen receptor (ER) levels (Foekens *et al.* 2000). The dataset follows preprocessing methods from (Royston and Altman 2013).
3) METABRIC: This breast cancer dataset includes gene expressions (*EGFR*, *PGR*, *ERBB2*, *MKI67*) and clinical features (age, treatments, estrogen receptor status) for 1523 training and 381 testing patients (Katzman *et al.* 2018).
4) FLCHAIN: This dataset includes 7874 subjects and examines the association between immunoglobulin light chain concentration and mortality. Key covariates are age, sex, year of sample, serum free light chain levels (kappa and lambda), FLC group, serum creatinine, and monoclonal gammopathy (MGUS) status (Dispenzieri *et al.* 2012). The dataset is randomly divided into an 80% training set and a 20% testing set.
5) NWTCO: The National Wilm's Tumor Study dataset includes 4028 children from the 3rd and 4th clinical trials, specifically focused on relapse events in Wilms' tumor. Covariates include histology assessments, cancer stage, clinical trial number, age, and a subcohort indicator (Breslow and Chatterjee 1999). The dataset is randomly divided into 80% for training and 20% for testing.

## 3.3 Evaluation on genomics data

Next, we evaluated CoxKAN on high-dimensional genomics datasets derived from The Cancer Genome Atlas Program (TCGA). These datasets include Copy Number Variations (CNVs) which reflect the mean amplification or deletion of genes or chromosomal regions relative to a reference genome; mRNA expression, which reflects gene expression levels and is derived from RNA sequencing; and binary indicators of mutation status for various genes. These datasets are characterized by high dimensionality and low sample size, with each type of feature presenting unique challenges. (i) CNVs: These features tend to show multicollinearity, which can complicate the interpretation of model coefficients and affect model stability. (ii) mRNA expression: These features often exhibit heavily skewed distributions, which can complicate statistical analysis and modeling. (iii) Mutation status: This type of feature is characterized by sparsity, with most entries being 0, which can make it challenging to detect meaningful patterns. These characteristics significantly increase the risk of overfitting, thus providing a "stress test" for CoxKAN.

In total, we curated five genomics datasets with diverse cancer types: breast invasive carcinoma (BRCA), stomach adenocarcinoma (STAD), glioma (GBM/LGG), lung adenocarcinoma (LUAD), and kidney renal clear cell carcinoma (KIRC). To ensure a representative test set, we divided each dataset into training (80%) and test (20%) sets by stratifying according to the distribution of observed durations and event indicators. All datasets include sparse mutation features and heavily skewed mRNA expression data. The GBM/LGG and KIRC datasets, as preprocessed in (Chen *et al.* 2022), also exhibit significant multicollinearity in the CNV features. For STAD, LUAD, and BRCA, we preprocessed the datasets to solve the multicollinearity issue in the features. Specifically, the preprocessing pipeline of STAD and BRCA involves: (i) Features were selected based on *P*-values derived from univariate CoxPH analysis. (ii) Groups of highly correlated features were consolidated by replacing them with a single feature representing the median value. (iii) Missing values were imputed by the random forest imputation method.

As a result, the BRCA dataset contains 811 training patients, 205 testing patients, and has 168 features in total (73 CNVs, 91 RNAs, 4 Mutations). The STAD dataset contains 284 training patients, 71 testing patients, and has 148 features (67 CNVs, 61 RNAs, and 20 Mutations). The LUAD dataset contained 289 training patients and 97 test patients, each with 429 RNA features. The GBM/LGG dataset contains 400 training patients and 100 testing patients. There are 320 features in total, consisting of the mutation status of the *IDH1* gene, 240 RNAs, and 79 CNVs (including the binary status of 1p19q arm codeletion). Finally, the KIRC dataset contains 388 training patients, 97 testing patients, and consists of 362 features (116 CNVs, 240 RNAs, and 6 Mutations).

# 4 Results

## 4.1 Synthetic datasets

The prediction results (C-Index) for each dataset are shown in Table 1, with pruned CoxKANs (pre-symbolic fitting) visualized in Fig. 2a. Pruning in CoxKAN successfully removed the irrelevant features, demonstrating automatic feature selection and yielding a shape well-suited to each problem. In contrast, CoxPH fails to predict the hazard function accurately in all cases. Note that since survival time is a random variable, the true formula does not achieve C-Index = 1. In the limit of an infinite dataset, achieving a higher C-Index than the true formula would be impossible.

CoxKAN successfully recovered three of the four hazard functions. The learned CoxKAN formulae are detailed in the Appendix: Table 1, available as supplementary data at *Bioinformatics* online. Specifically, for the "Gaussian" dataset, the CoxKAN formula nearly replicates the true formula. The formulas for the "Mixed" and "Euclidean" datasets were derived from the CoxKAN formulae through transformation and approximation, as shown in Appendix, available as supplementary data at *Bioinformatics* online (Formulae Transformation). Although the formula for the "Complex" dataset varies from the original function, Table 1 indicates no statistically significant differences in C-Index. Additionally, the distributions of the true and learned formulae visualized in Fig. 2b are nearly identical, demonstrating that CoxKAN provides a robust approximation.

## 4.2 Clinical datasets

Of the three clinical datasets, three of them (SUPPORT, GBSG, and METABRIC) were sourced from DeepSurv (Katzman *et al.* 2018), allowing us to compare directly with published DeepSurv results. We report the performance of CoxKAN at three stages: after training ("CoxKAN Trained"), after pruning ("CoxKAN Pruned"), and following symbolic fitting ("CoxKAN Symbolic").

The comparison results are shown in Table 2. We can see that CoxKAN Symbolic outperforms both CoxPH and DeepSurv across all datasets except FLCHAIN. Specifically, CoxKAN Symbolic shows an improvement in C-Index of up to 4.07% over CoxPH and 2.40% over DeepSurv, demonstrating consistent superior performance. Although FLCHAIN is the exception, CoxKAN Symbolic remains competitive, achieving a C-Index close to the best-performing method, CoxPH, with a difference of only 0.16%. Notably, CoxKAN Symbolic achieves a higher C-Index than CoxKAN Trained on 4 of 5 datasets, though the difference is not statistically significant. This supports the idea that pruning and symbolic fitting reduce variance error by eliminating irrelevant features and smoothing activations, introducing an inductive bias toward simpler functions that enhances generalization.

**Table 1.** The performance of CoxKAN and CoxPH over four synthetic datasets: C-index (95% confidence interval).

| Dataset | True formula | CoxPH | CoxKAN symbolic |
|---|---|---|---|
| Gaussian | $0.760_{(0.759,0.760)}$ | $0.499_{(0.497,0.500)}$ | $0.760_{(0.758,0.760)}$ |
| Mixed | $0.760_{(0.759,0.761)}$ | $0.688_{(0.688,0.690)}$ | $0.760_{(0.759,0.761)}$ |
| Euclidean | $0.725_{(0.724,0.726)}$ | $0.511_{(0.510,0.513)}$ | $0.723_{(0.721,0.723)}$ |
| Complex | $0.690_{(0.689,0.691)}$ | $0.664_{(0.663,0.665)}$ | $0.690_{(0.689,0.691)}$ |

The CoxKAN formulae learned for each dataset are provided in Appendix, available as supplementary data at *Bioinformatics* online [CoxKAN Formulae (Clinical)], with the corresponding networks and non-linear terms displayed in Figs. 2d and 3, respectively.

### 4.2.1 SUPPORT

The resulting CoxKAN network, which has three hidden neurons, can be separated into two main sub-networks, the full expression is given in Appendix: Equation (1), available as supplementary data at *Bioinformatics* online. The first sub-network, involving the first and third hidden neurons, has linear activations in the second layer that effectively skips a layer, indicating independent contributions to the log-partial hazard. We apply standard auto-symbolic fitting to this sub-network.

The second sub-network, involving the second hidden neuron, captures a complex interaction between patient age and cancer status. The second layer activation function in this sub-network, represented by $\phi_{1,1,2}$, is particularly intricate, and thus no symbolic operator from the default `auto_symbolic` method can effectively capture the variable interaction it encodes. To achieve a precise symbolic representation, we use PySR [Appendix: Equation (2), available as supplementary data at *Bioinformatics* online]. This case demonstrates how `auto_symbolic` may compromise key information, while the PySR expression retains it, as shown in Fig. 2c.

The second sub-network is visualized in Fig. 2d, where each data point represents the activation for a patient. The interaction term $\phi_{1,1,2}$ is re-plotted, and color-coded by cancer status and age. The results reveal the following insights:

- Patients with non-metastatic cancer show a high initial risk that decreases with age until 60, then rises.
- Patients without cancer have a lower risk that linearly increases with age.
- Patients with metastatic cancer are at the highest risk, increasing non-linearly with age.

### 4.2.2 GBSG

The CoxKAN log-partial hazard formula is provided in Appendix Equation (3), available as supplementary data at *Bioinformatics* online. A single trough observed in the activation functions for age and PGR suggests an optimal range, or "sweet spot", for these covariates. Key insights from this dataset include an increased patient risk associated with larger tumor size, higher lymph node count, and menopausal status, while hormonal therapy and higher estrogen receptor (ER) concentration contribute to lower risk.

### 4.2.3 METABRIC

The full CoxKAN expression for this dataset is presented in Appendix: Equation (4), available as supplementary data at *Bioinformatics* online. The effects of gene expressions in breast cancer are well-documented, with *PGR* linked to a favorable prognosis, while *ERBB2* and *MKI67* are associated with more aggressive tumors and poorer outcomes (Harris *et al.*, Harris *et al.*). CoxKAN successfully rediscovers these associations, providing precise symbolic formulae that align with established findings.

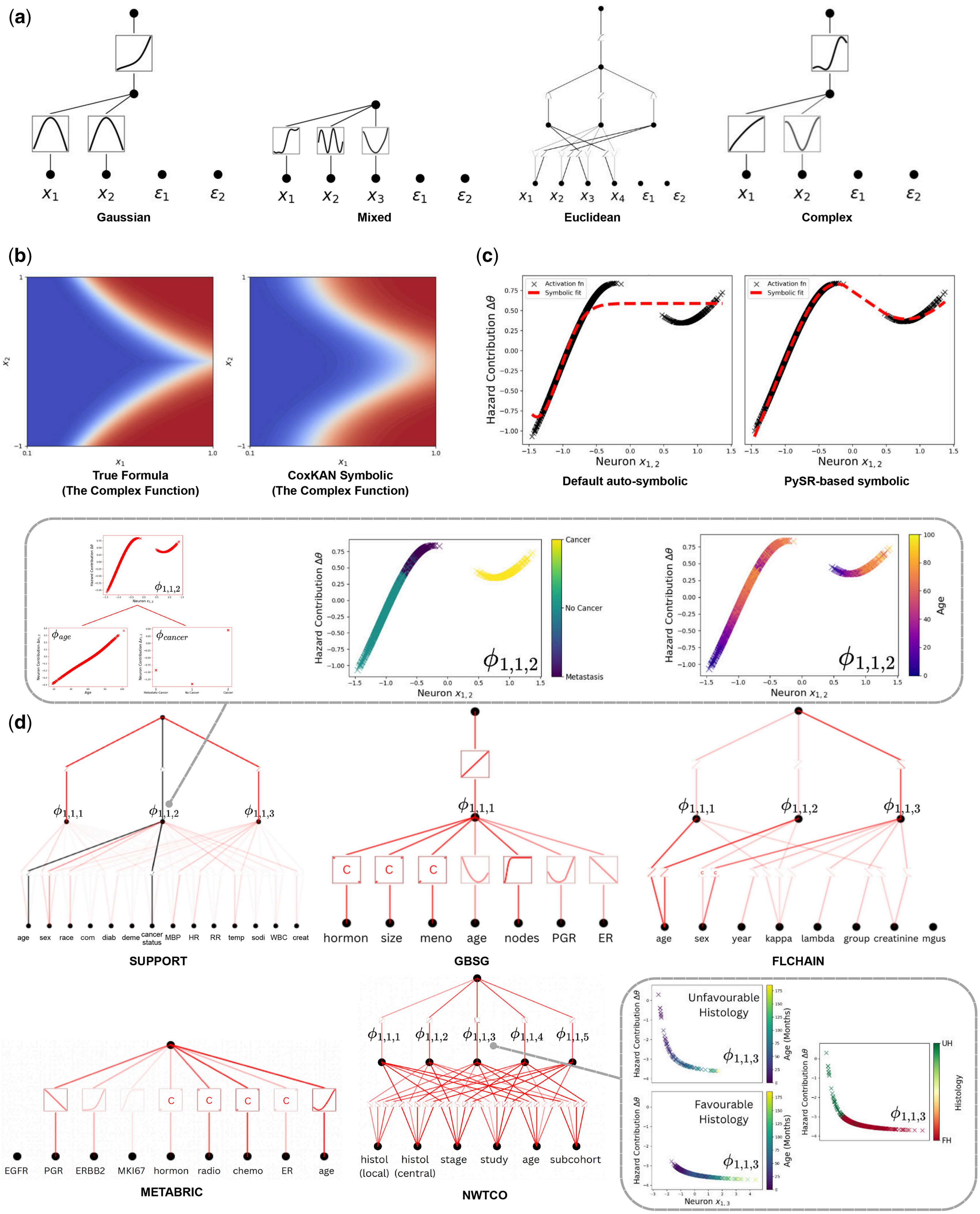

**Figure 2.** Visualization Overview. (a) Pruned CoxKAN networks for four synthetic datasets. Each $\varepsilon$ term represents an irrelevant feature, effectively pruned in all cases. (b) Log-partial hazard surfaces of the "Complex" dataset's true versus CoxKAN-predicted formula. (c) Symbolic fitting comparison for a complex activation function, using default auto-symbolic versus PySR-based symbolic fitting. (d) Trained CoxKAN networks for five clinical datasets. For SUPPORT, the interaction sub-network between age and cancer status is highlighted, with $\phi_{1,1,2}$ colored by cancer status and age, respectively. For NWTCO, $\phi_{1,1,3}$ interaction terms are shown with different colors, indicating central histology readings, and age-based colorations for patients with favorable or unfavorable histology, respectively.

**Table 2.** The benchmarking performance of CoxKAN and various models on five clinical datasets: C-Index (95% Confidence Interval).[a]

| Dataset | CoxPH | DeepSurv | SuMo | Cox-nnet | CoxKAN trained | CoxKAN pruned | CoxKAN symbolic |
|---|---|---|---|---|---|---|---|
| SUPPORT | $0.583_{(0.581,0.585)}$ | $0.618^{b}$ $_{(0.616,0.620)}$ | $0.601_{(0.600,0.604)}$ | $0.601_{(0.598,0.602)}$ | $\mathbf{0.624}_{(0.622,0.625)}$ | $\mathbf{0.624}_{(0.622,0.625)}$ | $0.623_{(0.623,0.626)}$ |
| GBSG | $0.656_{(0.655,0.662)}$ | $0.668^{b}$ $_{(0.665,0.671)}$ | $0.677_{(0.675,0.681)}$ | $0.673_{(0.669,0.675)}$ | $0.678_{(0.676,0.682)}$ | $0.679_{(0.675,0.681)}$ | $\mathbf{0.683}_{(0.678,0.684)}$ |
| METABRIC | $0.632_{(0.628,0.637)}$ | $0.643^{b}$ $_{(0.640,0.648)}$ | $0.626_{(0.622,0.630)}$ | $0.640_{(0.639,0.646)}$ | $0.647_{(0.644,0.652)}$ | $0.648_{(0.646,0.654)}$ | $\mathbf{0.650}_{(0.644,0.651)}$ |
| FLCHAIN | $\mathbf{0.798}_{(0.797,0.802)}$ | $0.795_{(0.793,0.798)}$ | $0.794_{(0.793,0.798)}$ | $0.796_{(0.794,0.799)}$ | $0.797_{(0.796,0.801)}$ | $0.796_{(0.792,0.797)}$ | $0.796_{(0.795,0.800)}$ |
| NWTCO | $0.698_{(0.693,0.703)}$ | $0.699_{(0.694,0.704)}$ | $0.703_{(0.698,0.709)}$ | $0.699_{(0.693,0.704)}$ | $0.720_{(0.714,0.725)}$ | $0.721_{(0.708,0.718)}$ | $\mathbf{0.722}_{(0.715,0.725)}$ |
| **Average** | 0.673 | 0.685 | 0.680 | 0.682 | 0.689 | 0.694 | 0.695 |

[a] Highest C-Index in bold.
[b] DeepSurv results on SUPPORT, GBSG, and METABRIC are quoted from the official DeepSurv publication.

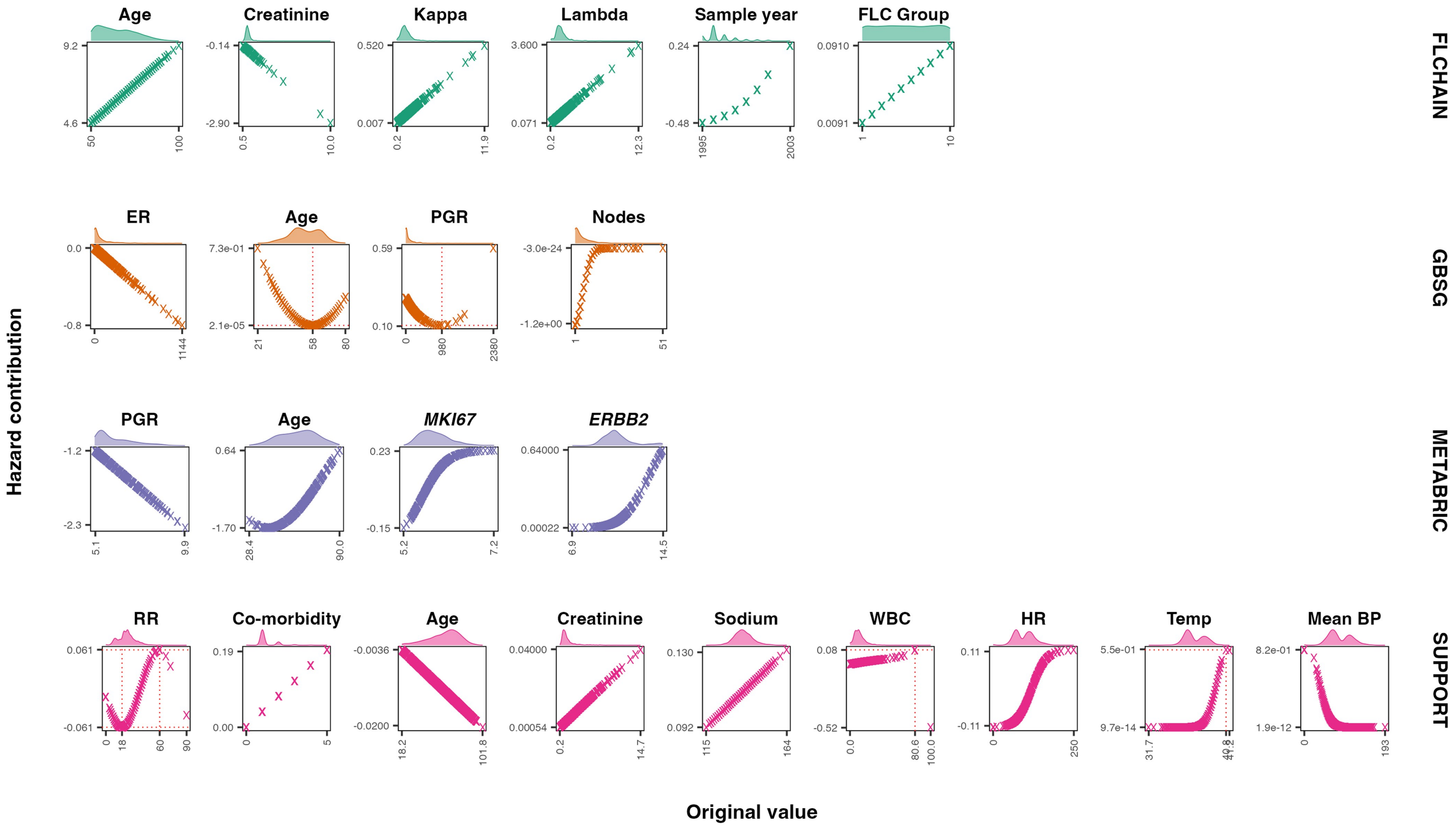

**Figure 3.** Non-linear hazard contributions for each variable in CoxKAN-generated equations for clinical datasets. Dotted lines mark the local maxima and minima, indicating shifts in hazard contribution. Density plots above each facet illustrate the distribution of original values for each non-linear term.

### 4.2.4 FLCHAIN

The CoxKAN formula for this dataset is shown in Appendix: Equation (5), available as supplementary data at *Bioinformatics* online, revealing several risk factors. Male sex, increasing age, and free light chains are associated with higher risk, along with FLC grouping, which adds a minor effect. Additionally, elevated serum creatinine levels are linked to increased risk, highlighting the importance of kidney function.

### 4.2.5 NWTCO

The learned CoxKAN network for this dataset comprises five hidden neurons. Linear activations, $\phi_{1,1,1}$, $\phi_{1,1,2}$, and $\phi_{1,1,5}$, create isolation terms, while non-linear activations, $\phi_{1,1,3}$ and $\phi_{1,1,4}$, introduce interactions [Appendix: Equation (6), available as supplementary data at *Bioinformatics* online]. The significant interaction term $\phi_{1,1,3}$ is detailed in Appendix: Equation (7), available as supplementary data at *Bioinformatics* online. From the isolation terms, we find that unfavorable histology and advanced cancer stages correlate with poorer prognosis. The interaction term $\phi_{1,1,3}$ (Fig. 2d) reveals:

- For patients with favorable histology, the prognosis improves with age.
- For unfavorable histology, prognosis worsens with increasing age, especially among younger patients.

## 4.3 Genomics datasets

Similarly to the previous section, we compare the performance of CoxKAN to CoxPH and DeepSurv. To ensure a fair comparison and address multicollinearity issues, we evaluated CoxPH with heavy Lasso regularization ("CoxPH Lasso").

The results are shown in Table 3. It is clear that CoxPH without regularization either encounters numerical problems or is only slightly better than random guessing. Introducing heavy Lasso regularization significantly improves the performance of CoxPH, even outperforming DeepSurv to a statistically significant degree on the STAD dataset. CoxKAN Symbolic demonstrates consistent and robust performance; it is either competitive with or surpasses CoxPH Lasso and DeepSurv on all datasets.

The CoxKAN formulae learned for each dataset are provided in Appendix, available as supplementary data at *Bioinformatics* online [CoxKAN Formulae (Genomics)]. Given the high dimensionality of features in these datasets, the log-partial hazard formulae derived using CoxKAN become quite large. To simplify these formulae, we estimate the relative importance of each term using its standard deviation $\sigma$ over the full dataset. Terms with higher standard deviations have a greater impact on the log-partial hazard, thus we only present the terms with high standard deviation ($\sigma$). We examine the interpretable log-partial hazard formulae produced by CoxKAN on the GBM/LGG and BRCA datasets, where CoxKAN Symbolic outperforms CoxPH with Lasso regularization, offering substantial new insights. For STAD and KIRC, CoxKAN Symbolic achieves performance comparable to CoxPH with Lasso regularization but does not add significant interpretive value beyond what CoxPH provides.

### 4.3.1 GBM/LGG

The full CoxKAN expression for the GBM/LGG dataset is found in Appendix: Equation (9), available as supplementary data at *Bioinformatics* online, with the non-linear terms shown in Fig. 4. 1p/19q co-deletion and *IDH1* mutation both have a linear, negative contribution to the hazard. Deletions of tumor suppressors *CDKN2A/B* have a positive contribution to the hazard. 10q CNV exhibits a non-linear impact on the hazard. An increase in *PTEN* and *JAK2* CNV has a positive contribution to the hazard, while *EGFR* amplification negatively contributes to the hazard. Any CNV in *CARD11* increases the hazard contribution. The role of many of these genes in glioma development progression has already been reported [1p/19q co-deletion: (Ostrom *et al.* 2014), *IDH1*: (Network 2015), *CDKN2A/B*: (Komori 2022), *PTEN*: (Wemmert *et al.* 2005, Ni *et al.* 2022), *EGFR*: (Zhao *et al.* 2021)]. However, there are currently no studies indicating a role for *CARD11* and *JAK2* in glioma progression. Our findings suggest that further research is needed to understand their biological function.

**Table 3.** The benchmarking performance of CoxKAN and various models on five genomic datasets: C-Index (95% Confidence Interval).[a]

| Dataset | CoxPH | CoxPH Lasso | DeepSurv | SuMo | Cox-nnet | CoxKAN trained | CoxKAN pruned | CoxKAN symbolic |
|---|---|---|---|---|---|---|---|---|
| GBM/LGG | N/A[b] | 0.788 (0.777, 0.799) | 0.819 (0.820, 0.836) | 0.813 (0.813, 0.829) | 0.791 (0.788, 0.806) | 0.814 (0.801, 0.816) | 0.811 (0.804, 0.820) | **0.818** (0.817, 0.828) |
| BRCA | 0.540 (0.529, 0.560) | 0.613 (0.607, 0.635) | 0.631 (0.622, 0.647) | 0.626 (0.608, 0.629) | 0.613 (0.597, 0.625) | 0.619 (0.593, 0.621) | **0.635** (0.617, 0.638) | 0.630 (0.622, 0.642) |
| STAD | 0.543 (0.521, 0.543) | 0.677 (0.673, 0.694) | 0.645 (0.629, 0.654) | 0.562 (0.552, 0.576) | 0.603 (0.592, 0.613) | **0.700** (0.697, 0.715) | 0.670 (0.665, 0.690) | 0.671 (0.659, 0.682) |
| LUAD | N/A[b] | 0.581 (0.566, 0.589) | 0.593 (0.582, 0.606) | **0.594** (0.570, 0.592) | 0.498 (0.485, 0.518) | 0.565 (0.549, 0.574) | 0.554 (0.540, 0.569) | 0.553 (0.549, 0.578) |
| KIRC | N/A[b] | 0.686 (0.663, 0.686) | 0.636 (0.626, 0.651) | 0.663 (0.662, 0.683) | **0.691** (0.682, 0.703) | 0.672 (0.662, 0.684) | 0.672 (0.668, 0.688) | 0.668 (0.669, 0.689) |
| **Average** | N/A[b] | 0.668 | 0.665 | 0.652 | 0.642 | 0.674 | 0.668 | 0.668 |

[a] Highest C-Index in bold.
[b] In the presence of multicollinearity, the design matrix is non-invertible and CoxPH fails without regularization.

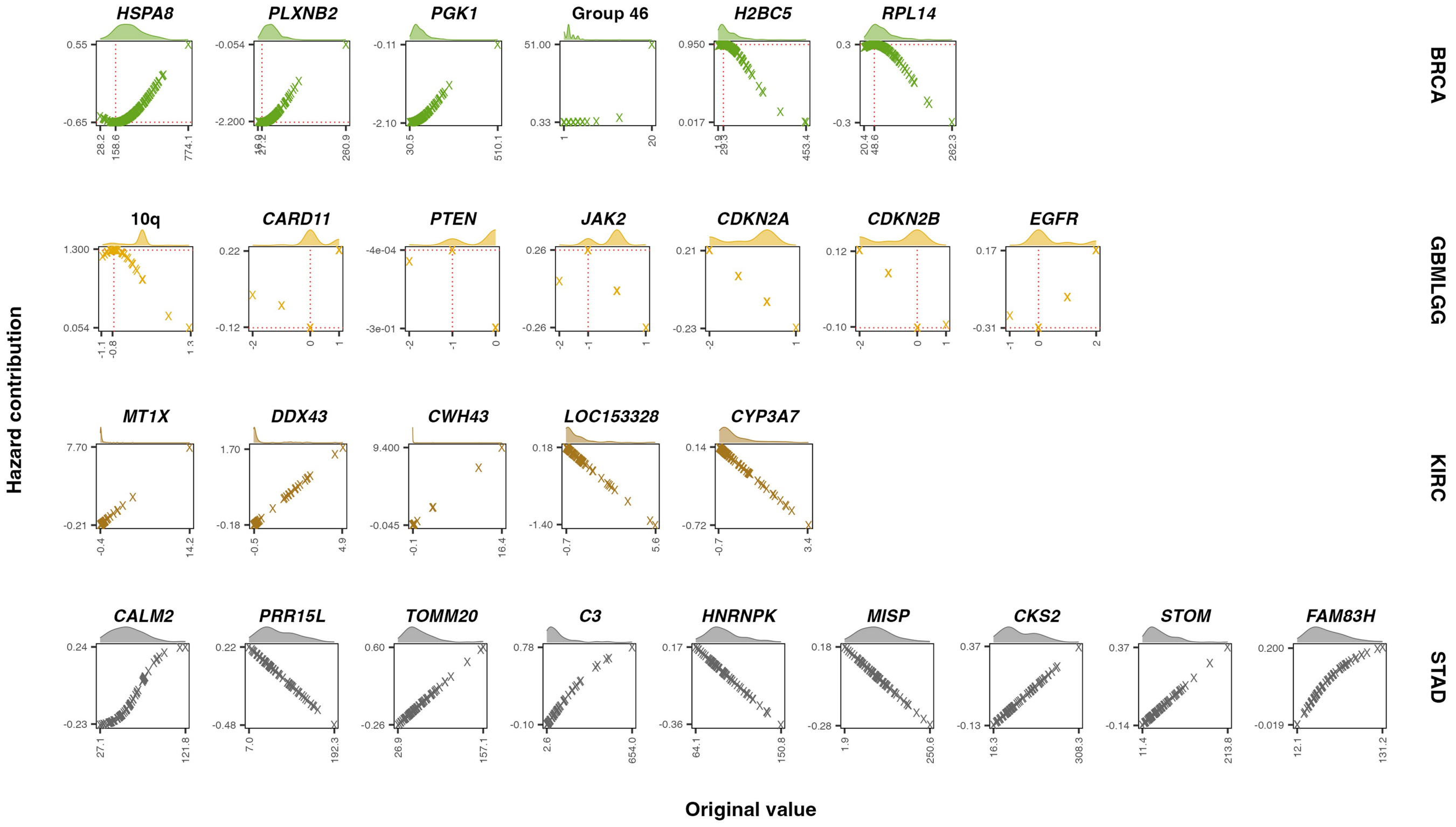

**Figure 4.** The hazard contribution for non-linear terms in the CoxKAN-generated equations for the genomic datasets. The LUAD dataset is omitted due to a lack of non-linear terms. Dotted lines mark the local maxima and minima, indicating shifts in hazard contribution. Density plots above each facet illustrate the distribution of original values for each non-linear term.

#### 4.3.2 BRCA

The full CoxKAN expression for the BRCA dataset is found in Appendix: Equation (10), available as supplementary data at *Bioinformatics* online, with the non-linear terms shown in Fig. 4. Mutations in *KMT2C*, *DMD*, *TTN* have a linear, positive association with the hazard. *RYR2* has a linear, negative association with the hazard for breast cancer patients. This is reflective of breast cancer associations reported in the literature [*KMT2C*: (Tinsley *et al.* 2024), *DMD*: (Luce *et al.* 2017), *TTN*: (Agarwal *et al.* 2024), *RYR2*: (Xu *et al.* 2021)]. The mRNA expression of *PLXNB2*, *PGK1*, *H2BC5*, and Group 46 CNV, exhibit non-linear, monotonic relationships with the hazard. *HSPA8* and *RPL14* expressions exhibit mostly monotonic behavior across their ranges. Notably, *HSPA8* and *RPL14* show reverse effects in certain small ranges, indicating that both over-expression and under-expression can influence patient risk. These results highlight the need for more research to better understand these genes' biological roles.

#### 4.3.3 LUAD

The full CoxKAN expression for the LUAD dataset is found in Appendix Equation (13), available as supplementary data at *Bioinformatics* online. All the relationships learned between the hazard and mRNA expression are linear. The mRNA expression in *CLN6* and *DHRS1* have the strongest negative association with the hazard, *TFPI2* and *DCP2* have the strongest positive association with the hazard. For *TFPI2* (Rollin *et al.* 2005, Wu *et al.* 2012) and *DCP2* (Wang *et al.* 2022), this replicates other findings in the literature. However, no specific connection between *CLN6* and survival of lung adenocarcinoma has been previously found, although a general connection between lysosomal gene expression and survival was established in (Li *et al.* 2023). Further, a connection between *DHRS1* has been established in hepatocellular carcinoma, but not lung adenocarcinoma specifically (Guo *et al.* 2021). Our results provide evidence of novel, sensible associations between *CLN6* and *DHRS1* expression and lung adenocarcinoma survival (Figs. 2–4).

## 5 Discussion

This paper presented the CoxKAN framework, a novel application of Kolmogorov-Arnold Networks to interpretable survival regression. We demonstrated that CoxKAN achieves (i) sophisticated interpretability by obtaining symbolic formulae for the hazard function and visualizing KAN activation functions and (ii) high performance due to the ability to flexibly capture any function (low bias error). We were also able to mitigate CoxKAN overfitting, which can be attributed to the explicit regularization in the loss function, early stopping, and the inductive bias of the pruning and symbolic fitting pipeline that encourages simpler functions, which generalize better than the original network.

In the first series of experiments, we generated synthetic datasets using custom symbolic formulae for the hazard function and found that in 3/4 examples CoxKAN was able to recover the symbolic form. In the last example (which was intentionally difficult to recover), CoxKAN found an accurate approximation to the ground truth; we claim that CoxKAN still possesses the properties of interpretability and high performance in this case. Additionally, CoxKAN automatically pruned the irrelevant, noisy features added to all synthetic datasets, demonstrating successful feature selection. We then evaluated CoxKAN on five clinical datasets and five high-dimensional genomics datasets. On the clinical data, CoxKAN Symbolic achieved a statistically significant improvement in performance over all other methods in 3/5 datasets. On the genomics data, CoxKAN Symbolic achieved a statistically significant performance improvement over all others methods in 2/5 datasets. On datasets that CoxKAN did not outperform other methods, the performance difference was generally not statistically significant as characterized by overlapping confidence intervals with one exception; Cox-nnet significantly outperform CoxKAN on TCGA-KIRC. CoxKAN also uncovered useful insights from the survival data. For example, on the SUPPORT dataset, CoxKAN identified that the risk of cancer patients in metastasis decreases with age until about 60 years old, then starts to increase, but for patients with non-metastatic cancer or no cancer at all, their risk only increases with age. This kind of variable interaction would be extremely difficult to identify using existing survival models. On the genomics datasets, CoxKAN uncovered a number of important biological associations between cancer risk and genomic features such as specific CNVs and mRNA transcripts, offering valuable insights that can guide further biological studies and the development of targeted therapeutic strategies.

Our work has several limitations. Firstly, CoxKAN is exposed to the bias of the proportional-hazards assumptions such as covariate time independence and the baseline hazard being consistent for all patients. An exciting future direction would be to construct a KAN-based framework that bypasses these assumptions while retaining precise interpretability. Secondly, CoxKAN is vulnerable to overfitting and thus for the high dimensional genomics datasets, the hyperparameter search typically yielded low-capacity KANs with no hidden layers. This meant that interactions between genomic features were not learned, even though it is well known that genomics features do experience interactions. Additional effort to mitigate overfitting while retaining the ability to capture interactions is a promising future direction. Furthermore, the performance CoxKAN tended to be unstable with respect to initialization, and training was sensitive to hyper-parameters. These flaws could be addressed by experimenting with more techniques related to hyperparameter tuning, regularization, and optimization.

In conclusion, CoxKAN offers a framework for survival analysis that is precisely interpretable, high-performing, and offers the full flexibility of a deep learning environment. Its applications are far ranging, from biomarker discovery to treatment efficacy studies; and could even be applied beyond the medical field, for engineering, business, or finance time-to-event analyses.

## Acknowledgements

Calculations were performed in part using the Sulis Tier 2 HPC platform hosted by the Scientific Computing Research Technology Platform at the University of Warwick. Sulis is funded by EPSRC Grant EP/T022108/1 and the HPC Midlands+ consortium. The funders had no role in study design, data collection and analysis, decision to publish, or

## Author contributions

William Knottenbelt (Conceptualization [equal], Data curation [equal], Investigation [lead], Methodology [lead], Software [lead], Validation [lead], Visualization [equal], Writing—original draft [equal], Writing—review & editing [equal]), William McGough (Data curation [equal], Formal analysis [equal], Investigation [equal], Methodology [equal], Software [equal]), Rebecca Wray (Formal analysis [equal], Visualization [lead], Writing—review & editing [equal]), Woody Zhidong Zhang (Formal analysis [equal], Writing—review & editing [equal]), Jiashuai Liu (Data curation [equal], Writing—review & editing [equal]), Ines Prata Machado (Data curation [equal], Investigation [equal], Writing—review & editing [equal]), Zeyu Gao (Conceptualization [lead], Data curation [lead], Formal analysis [equal], Investigation [equal], Methodology [equal], Resources [equal], Supervision [equal], Writing—original draft [lead]), and Mireia Crispin-Ortuzar (Conceptualization [equal], Funding acquisition [lead], Project administration [lead], Supervision [equal], Writing—review & editing [lead])

## Supplementary data

Supplementary data is available at *Bioinformatics* online.

Conflict of interest: Z.G.: GE HealthCare (research funding). M.C.O.: 52 North Health Ltd (co-founder and employee), GE HealthCare (research funding), GSK (speaking fees). The remaining authors declare no competing interests.

## Funding

We acknowledge funding and support from Cancer Research UK and the Cancer Research UK Cambridge Centre [CTRQQR-2021–100012], The Mark Foundation for Cancer Research [RG95043], GE HealthCare, and the CRUK National Cancer Imaging Translational Accelerator (NCITA) [A27066]. This work was supported by the Cancer Research UK Cambridge Centre [CTRQQR-2021/100012] and the International Alliance for Cancer Early Detection, a partnership between Cancer Research UK [C14478/A27855], Canary Center at Stanford University, the University of Cambridge, OHSU Knight Cancer Institute, University College London and the University of Manchester. Additional support was also provided by the National Institute of Health Research (NIHR) Cambridge Biomedical Research Centre [NIHR203312] and EPSRC Tier-2 capital grant [EP/P020259/1].

## Data availability

All datasets used in this study were obtained from publicly accessible sources. Specifically, clinical datasets were retrieved from the following repositories: GBSG (https://www.kaggle.com/datasets/utkarshx27/breast-cancer-dataset-used-royston-and-altman), METABRIC (https://www.kaggle.com/datasets/raghadalharbi/breast-cancer-gene-expression-profiles-metabric), SUPPORT (https://hbiostat.org/data), NWTCO and FLCHAIN (https://github.com/vincentarelbundock/Rdatasets). In addition, genomics datasets for STAD, BRCA, GBM/LGG, KIRC, and LUAD were obtained from The Cancer Genome Atlas (TCGA) portal (https://portal.gdc.cancer.gov).

## References

Agarwal S, Parija M, Naik S *et al.* Dysregulated gene subnetworks in breast invasive carcinoma reveal novel tumor suppressor genes. *Sci Rep* 2024;**14**:15691.

Breslow NE, Chatterjee N. Design and analysis of two-phase studies with binary outcome applied to WILMS tumour prognosis. *J R Stat Soc Ser C (Appl Stat)* 1999;**48**:457–68.

Chen RJ, Lu MY, Wang J *et al.* Pathomic fusion: an integrated framework for fusing histopathology and genomic features for cancer diagnosis and prognosis. *IEEE Trans Med Imaging* 2022;**41**:757–70.

Cheon M. Kolmogorov-Arnold network for satellite image classification in remote sensing. arXiv preprint arXiv:2406.00600, 2024.

Ching T, Zhu X, Garmire LX. Cox-nnet: an artificial neural network method for prognosis prediction of high-throughput omics data. *PLoS Comput Biol* 2018;**14**:e1006076.

Cox DR. Regression models and life-tables. *J R Stat Soc Ser B (Methodological)* 1972;**34**:187–202.

Cranmer M. Interpretable machine learning for science with PYSR and symbolicregression.jl. arXiv preprint arXiv:2305.01582, 2023.

Dispenzieri A, Katzmann JA, Kyle RA *et al.* Use of nonclonal serum immunoglobulin free light chains to predict overall survival in the general population. *Mayo Clin Proc* 2012;**87**:517–23.

Faraggi D, Simon R. A neural network model for survival data. *Stat Med* 1995;**14**:73–82.

Fernandez T, Rivera N, Teh YW. Gaussian processes for survival analysis. *Adv Neural Inf Process Syst* 2016;**29**:5021–9.

Foekens JA, Peters HA, Look MP *et al.* The urokinase system of plasminogen activation and prognosis in 2780 breast cancer patients. *Cancer Res* 2000;**60**:636–43.

Guo D-Z, Huang A, Wang Y-P *et al.* Development of an eight-gene prognostic model for overall survival prediction in patients with hepatocellular carcinoma. *J Clin Transl Hepatol* 2021;**9**:898–908.

Harris L, Fritsche H, Mennel R *et al.*; American Society of Clinical Oncology 2007. Update of recommendations for the use of tumor markers in breast cancer. *J Clin Oncol* 2007;**25**:5287–312.

Ishwaran H, Kogalur UB. Random survival forests for r. *R News* 2007;**7**:25–31.

Ishwaran H, Kogalur UB, Blackstone EH *et al.* Random survival forests. *Ann Appl Stat* 2008;**2**:841–60.

Katzman JL, Shaham U, Cloninger A *et al.* Deepsurv: personalized treatment recommender system using a cox proportional hazards deep neural network. *BMC Med Res Methodol* 2018;**18**:24.

Kim S, Kim K, Choe J *et al.* Improved survival analysis by learning shared genomic information from pan-cancer data. *Bioinformatics* 2020;**36**:i389–98.

Kingma D, Ba J. Adam: A method for stochastic optimization. In: *International Conference on Learning Representations (ICLR)*. San Diego, CA, USA, 2015.

Knaus WA, Harrell FE, Lynn J *et al.* The support prognostic model: objective estimates of survival for seriously ill hospitalized adults. *Ann Intern Med* 1995;**122**:191–203.

Komori T. Grading of adult diffuse gliomas according to the 2021 who classification of tumors of the central nervous system. *Lab Invest* 2022;**102**:126–33.

Kvamme H, Borgan O. Continuous and discrete-time survival prediction with neural networks. *Lifetime Data Anal* 2021;**27**:710–36.

Kvamme H, Borgan Ø, Scheel I. Time-to-event prediction with neural networks and cox regression. *J Mach Learn Res* 2019;**20**:1–30.

Langbein SH, Krzyziński M, Spytek M *et al.* Interpretable machine learning for survival analysis. arXiv preprint arXiv:2403.10250, 2024.

Le-Rademacher J, Wang X. Time-to-event data: an overview and analysis considerations. *J Thorac Oncol* 2021;**16**:1067–74.

Lee C, Zame W, Yoon J *et al.* Deephit: A deep learning approach to survival analysis with competing risks. In: *Proceedings of the AAAI*

*conference on artificial intelligence, New Orleans, Louisiana, USA*. Vol. 32. Washington, DC, USA: AAAI Press, 2018.

Li C, Liu X, Li W *et al.* U-kan makes strong backbone for medical image segmentation and generation. In: *Proceedings of the AAAI Conference on Artificial Intelligence. Philadelphia, Pennsylvania, USA*. Vol. 39. Washington, DC, USA: AAAI Press, 2024.

Li H, Sha X, Wang W *et al.* Identification of lysosomal genes associated with prognosis in lung adenocarcinoma. *Transl Lung Cancer Res* 2023;**12**:1477–95.

Liu Z, Wang Y, Vaidya S *et al.* Kan: Kolmogorov-Arnold networks. arXiv preprint arXiv:2404.19756, 2024.

Luce LN, Abbate M, Cotignola J *et al.* Non-myogenic tumors display altered expression of dystrophin (DMD) and a high frequency of genetic alterations. *Oncotarget* 2017;**8**:145–55.

Monnickendam G, Zhu M, McKendrick J *et al.* Measuring survival benefit in health technology assessment in the presence of nonproportional hazards. *Value Health* 2019;**22**:431–8.

Nagpal C, Yadlowsky S, Rostamzadeh N *et al.* Deep cox mixtures for survival regression. In: *Machine Learning for Healthcare Conference*. Cambridge, MA, USA: PMLR, 2021, 674–708.

Network CGAR. Comprehensive, integrative genomic analysis of diffuse lower-grade gliomas. *N Engl J Med* 2015;**372**:2481–98.

Ni X, Wu W, Sun X *et al.* Interrogating glioma-m2 macrophage interactions identifies gal-9/tim-3 as a viable target against PTEN-null glioblastoma. *Sci Adv* 2022;**8**:eabl5165.

Ostrom QT, Bauchet L, Davis FG *et al.* The epidemiology of glioma in adults: a "state of the science" review. *Neuro Oncol* 2014; **16**:896–913.

Ou FS, Michiels S, Shyr Y *et al.* Biomarker discovery and validation: statistical considerations. *J Thorac Oncol* 2021;**16**:537–45.

Rindt D, Hu R, Steinsaltz D *et al.* Survival regression with proper scoring rules and monotonic neural networks. In: *International Conference on Artificial Intelligence and Statistics. Valencia, Spain*. Cambridge, MA, USA: PMLR, 2022, 1190–205.

Rollin J, Iochmann S, Blechet C *et al.* Expression and methylation status of tissue factor pathway inhibitor-2 gene in non-small-cell lung cancer. *Br J Cancer* 2005;**92**:775–83.

Royston P, Altman DG. External validation of a cox prognostic model: principles and methods. *BMC Med Res Methodol* 2013;**13**:33.

Saegusa T, Zhao Z, Ke H *et al.* Detecting survival-associated biomarkers from heterogeneous populations. *Sci Rep* 2021;**11**:3203.

Tinsley E, Bredin P, Toomey S *et al.* Kmt2c and kmt2d aberrations in breast cancer. *Trends Cancer* 2024;**10**:519–30.

Vaca-Rubio CJ, Blanco L, Pereira R *et al.* Kolmogorov-Arnold networks (KANS) for time series analysis. arXiv preprint arXiv:2405.08790, 2024.

Vaswani A, Shazeer N, Parmar N *et al.* Attention is all you need. In: *Advances in Neural Information Processing Systems 30*. California, USA: Curran Associates, Inc., 2017

Wang G, Zhao M, Li J *et al.* m7g-associated subtypes, tumor microenvironment, and validation of prognostic signature in lung adenocarcinoma. *Front Genet* 2022;**13**:954840.

Wang P, Li Y, Reddy CK. Machine learning for survival analysis: a survey. *ACM Comput Surv* 2019;**51**:1–36.

Watanabe S. Tree-structured parzen estimator: understanding its algorithm components and their roles for better empirical performance. arXiv preprint arXiv:2304.11127, 2023.

Wemmert S, Ketter R, Rahnenfuhrer J *et al.* Patients with high-grade gliomas harboring deletions of chromosomes 9p and 10q benefit from temozolomide treatment. *Neoplasia* 2005;**7**:883–93.

Wiegrebe S, Kopper P, Sonabend R *et al.* Deep learning for survival analysis: a review. *Artif Intell Rev* 2024;**57**:65.

Wu D, Xiong L, Wu S *et al.* Tfpi-2 methylation predicts poor prognosis in non-small cell lung cancer. *Lung Cancer* 2012;**76**:106–11.

Xu Z, Xiang L, Wang R *et al.* Bioinformatic analysis of immune significance of ryr2 mutation in breast cancer. *Biomed Res Int* 2021; **2021**:8072796.

Yu C-N, Greiner R, Lin H-C *et al.* Learning patient-specific cancer survival distributions as a sequence of dependent regressors. In *Advances in Neural Information Processing Systems 24*. Granada, Spain: Curran Associates, Inc., 2011.

Zhao H-F, Zhou X-M, Wang J *et al.* Identification of prognostic values defined by copy number variation, mRNA and protein expression of lancl2 and EGFR in glioblastoma patients. *J Transl Med* 2021; **19**:372.